\documentclass[runningheads]{llncs}
\usepackage[numbers]{natbib}

\usepackage{times}  % DO NOT CHANGE THIS
\usepackage{helvet} % DO NOT CHANGE THIS
\usepackage{courier}  % DO NOT CHANGE THIS
\usepackage[hyphens]{url}  % DO NOT CHANGE THIS
\usepackage{graphicx} % DO NOT CHANGE THIS
\usepackage{graphicx}  % DO NOT CHANGE THIS
\usepackage{soul}
\usepackage{dirtytalk}
\usepackage{flushend}
\usepackage{multirow}

\usepackage{amsmath,amssymb,amsfonts}

\usepackage{textcomp}
\usepackage{xcolor}

\usepackage[export]{adjustbox}

\usepackage{subcaption}
\usepackage{comment}

\usepackage{tikz}
\usepackage{lipsum}

\usepackage{cancel}
\usetikzlibrary{automata,positioning,bending,arrows}
\tikzset{
->, % makes the edges directed
>=stealth, % makes the arrow heads bold
node distance=3cm, % specifies the minimum distance between two nodes. Change if necessary.
every state/.style={thick, fill=gray!10}, % sets the properties for each ’state’ node
initial text=$ $, % sets the text that appears on the start arrow
}

\usepackage[markup=underlined]{changes}
\definechangesauthor[color=red]{error}
\definechangesauthor[color=blue]{new}
\definechangesauthor[color=magenta]{anas}
\definechangesauthor[color=brown]{myquote}

\usepackage[textwidth=1.5cm]{todonotes}
\usepackage{cleveref}
\usepackage{changepage}

\begin{document}

\title{Goal-constrained planning domain model verification of safety properties\thanks{Supported by EPSRC grant EP/P510427/1 in collaboration with Schlumberger.}}

\titlerunning{Goal-constrained planning domain model verification of safety properties}

\author{Anas Shrinah \and
Kerstin Eder}
\authorrunning{A. Shrinah and K. Eder}
\institute{University of Bristol, Bristol BS8 1TH, UK 
\\ \email{\{first.last\}@bristol.ac.uk}}
\maketitle              %
\begin{abstract}
The verification of planning domain models is crucial to ensure the safety, integrity and correctness of planning-based automated systems. This task is usually performed using model checking techniques. 
However, unconstrained application of model checkers to verify planning domain models can result in false positives, i.e.\ counterexamples that are unreachable by a sound planner when using the domain under verification during a planning task. 
In this paper, we discuss the downside of unconstrained planning domain model verification. We then introduce the notion of a valid planning counterexample, and demonstrate how model checkers, as well as state trajectory constraints planning techniques, should be used to verify planning domain models so that invalid planning counterexamples are not returned. 
\end{abstract}

\section{Introduction}

Planning and task scheduling techniques are increasingly applied to real world problems such as activity sequencing, constraint solving and resource management. These processes are implemented in planning-based automated systems which are already used in space missions \cite{MUSCETTOLA19985,1373506,1265878}, search and rescue \cite{Hugh95knowledgelevel}, logistics \cite{tate1996plan} and many other domains. Since the failure of such systems could have catastrophic consequences, these applications are regarded as safety-critical. Therefore, verification methods that are robust, trustworthy and systematic are crucial to gain confidence in the safety, integrity and correctness of these systems. 

The literature is rich with studies on verification of planning systems. For instance, \citet{794352} carried out scenario-based testing and model-based validation of the remote agent that controlled the Deep Space 1 mission. Another example is the verification of the safety of the autonomous science agent design that was deployed on the Earth Orbiter 1 spacecraft~\cite{cichy2004validating}.

A typical planning system consists of a planning domain model, planning problem, planner, plan, executive, and monitor. Planners take as an input a domain model which describes application-specific states and actions, and a problem that specifies the planning goal and the initial state. From these inputs, a sequence of actions that can achieve the goal starting from the initial state is returned as plan. The plan is then executed by an executive to change the world state so that it matches the desired goal.

Our research focuses on the verification of planning domain models wrt.\ safety properties. Domain models provide the foundations for planning. They describe real-world actions by capturing their pre-conditio{}ns and effects.
Due to modelling errors, a domain model might be inconsistent, incomplete, or inaccurate~\cite{Bensalem2014}. This could cause the planner to fail in finding a plan or to generate unrealistic plans that will fail to execute in the real world. Moreover, erroneous domain models could lead planners to produce unsafe plans that, when executed, could have catastrophic consequences in the real world. 

This paper addresses the fact that the state-of-the-art verification methods for planning domain models are vulnerable to false positive counterexamples. In particular, unconstrained verification tasks might return counterexamples that are unreachable by planners when using the domain under verification (DUV) during a planning task. Such counterexamples can mislead designers to unnecessarily restrict domain models, thereby potentially blocking valid and possibly necessary behaviours.
In addition, false positive counterexamples can lead verification engineers to overlook counterexamples that are reachable by planners.
According to the Electronic Engineering Times, a leading technological news website in the electronics industry:

\begin{adjustwidth}{10pt}{10pt}
\say{When a design is under-constrained, illegal inputs can lead to the formal tool exploring illegal design states. The tool may report false bugs, resulting in the verification team spending time pursuing \say{wild-goose chases}. Under-constrained designs can also lead to a false sense of achieving the desired coverage.}~\cite{EETimes}
\end{adjustwidth}

This is a well studied problem in the Verification and Validation (V\&V) community. For instance, \citet{nguyen2008unbounded} mentioned that false counterexamples can be avoided by constraining the property with reachability information.
However, the literature, e.g. \citet{smith2005model}, suggests that this aspect has been overlooked in the context of planning domain model verification. We discuss this further in Section~\ref{Shortcomings}.
To address this oversight, we propose to employ planning goals as constraints during verification. 

Thus, we introduce {\em goal-constrained planning domain model verification}, a concept transferred from V\&V research that eliminates invalid planning counterexamples per se.
We formally prove that goal-constrained planning domain model verification of safety properties is guaranteed to return only valid planning counterexamples, if and only if any exist.
We also demonstrate two different ways to perform goal-constrained planning domain model verification, one using model checkers and the other using state trajectory constraints planning techniques. 
We illustrate our method using the Cave Diving planning domain model as an example~\cite{IPC2014}.
Additionally, we perform empirical experiments to demonstrate the feasibility and investigate the behaviour of our approach using the Spin model checker~\cite{holzmann2004spin} and the MIPS-XXL-IPC5 planner~\cite{edelkamp2006costoptimal}.
To the best of our knowledge, this work is the first to introduce the concept of {\em goal-constrained planning domain model verification}.

The rest of this paper is organised as follows. \Cref{related_work} contrasts the concepts presented here with related work.
\Cref{Shortcomings} informally discusses the problem of invalid planning counterexamples in planning domain model verification.
A verification concept of planning domain models that avoids returning invalid planning counterexamples is presented in \Cref{Verification_planning_domain_models}.
\Cref{Example_CaveDiving} discusses the application of this concept on our example domain.
\Cref{Experiments} reports and discusses the experimental results.
Finally, \Cref{conclusion} concludes the paper and suggests future work.

\section{Related Work} \label{related_work}
Closely related, but different, is the work by~\citet{Albarghouthi09onthe}.
Their main objective is to treat verification as a planning task, whereas our aim is to demonstrate how model checkers and planners can be used for domain model verification. 
They proposed to perform system model verification using classical planners. To do this, they first translated the model of the system under verification into a planning domain model. Then, the negation of the safety property to be established is used as the goal for the planner, which is then consulted to find a plan that acts as counterexample for the given property. 
In our study, because our aim is to verify domain models against a given safety property with respect to a specific goal,
we use state trajectory constraints to restrict counterexamples to identify plans that can achieve the planning goal while falsifying the safety property. 
In their work the negation of the safety property is used as the goal. Whereas, in our method, the negation of the safety property is represented as a state trajectory constraint and the goal is the given planning goal. 

\citet{raimondi2009pdver} also applied verification as planning to verify planning domain models, starting from LTL specifications~\cite{clarke1999model}.
This work fundamentally differs from our work. They tested the impact of individual atomic propositions on the validity of the overall verified property by translating the specification properties into trap formulas. However, their method does not consider the interaction between property testing and the original planning goal. 
Note that finding a planning constraint to exercise a specific atomic proposition is not enough to ensure the constraint itself would be exercised during the planning process. 
For example, a planning goal might be achieved through a state trajectory that does not exercise the hard constraint used to represent the tested property. Our work is mainly focused on investigating this interaction. Therefore, we use state trajectory constraints to guarantee the property is tested while achieving the planning goal.

\citet{goldman2012using} also used classical planners for planning systems verification, but they examined verifying plans rather than domain models. They proposed an approach that uses classical planners to find counterexamples for a given planning problem and plan instance. Their work and ours are related in that both suggest performing planning verification for a specific planning problem rather than attempting 
unconstrained verification of a planning system. However, their work is limited to the verification of single plan instances, whereas our method verifies all potential plans that can be spun from a domain model for a specific goal.

Among others, the researchers in \cite{penix1998using,khatib2000verification,smith2005model,havelund2008automated,cesta2010validation} used model checkers to verify planning domain models. 
They translated the respective domain models into the input language of the selected model checker. The model checker is then applied to verify the domain model wrt.\ a given specification property. 
Similarly, we also propose a method to verify domain models using model checkers. 
However, our method differs from the others in two aspects. First, in the way we define the planning domain model verification problem, and, second, in the way we use model checkers to perform verification. 
As explained in \Cref{Verification_planning_domain_models}, we constrain the verification of planning domain models with a specific goal. 
In contrast, previous studies perform unconstrained verification of domain models, i.e.\ they leave the goal open. 
As discussed in \Cref{Shortcomings}, the unconstrained goal may cause the model checker to return counterexamples that are unreachable when a planner uses the DUV. 
On the other hand, when the goal is constrained for verification, then we show that the returned counterexamples, if any, are guaranteed to be reachable by any sound planner. 
The second difference is that, after the planning domain model is translated to the model checker's input language, we augment the model transitions, introducing the negation of the goal as a new constraint, thereby forcing the model checker to terminate once the goal is reached.
This modification prevents the model checker from returning counterexamples that falsify the given property after satisfying the goal; these are unreachable by planners.

\section{Invalid counterexamples in planning domain model verification} \label{Shortcomings}

Planning domain model verification aims to demonstrate that any produced plan
satisfies a set of properties. 
To achieve this, formal planning domain model verification methods leave the planning goal open.
This, we define as {\em unconstrained} verification of planning domain models, i.e.\ the verification is expected to hold for any potential goal.
Unconstrained verification searches the domain model for a sequence of actions that can falsify the given property, regardless of any other conditions. In particular, whether or not a planner would consider this sequence to be a plan, is not taken into account. 
This is a critical oversight, because, when the domain model is used to solve a specific planning problem, the sequence of actions that constitutes such a counterexample might, in fact, be ``pruned away'' by the planner, if it does not satisfy the planning goal. 
Therefore, for a specific planning problem, counterexamples that do not achieve the planning goal are deemed unreachable counterexamples from the planner's perspective. Hence, we consider them to be invalid planning counterexamples.

\begin{figure}[ht] %
\centering %
\begin{tikzpicture}[shorten >=1pt,node distance=2.5cm,on grid,auto,scale=0.4, every node/.style={scale=0.7}] 
   
   \node[state,initial,label=below:{\small $s_0$}] (q_0) [align=center]   {!Close \\ !Start \\ !Error \\ !Heat};
   \node[state, label=below:{\small $s_1$}] (q_1) [align=center,above right=of q_0] {Close \\ !Start \\ !Error \\ !Heat}; 
   \node[state, label=below:{\small $s_2$}] (q_2) [align=center, below right=of q_0] {!Close \\ Start \\ Error \\ !Heat}; 
   \node[state,accepting,label=below:{\small $s_3$}](q_3) [align=center, below right=of q_1] {Close \\ Start \\ !Error \\ Heat};
    \path[->] 
     (q_0) edge  node [midway,above, sloped, align=center] {Close \\ Door} (q_1)
          edge  node [midway,above, sloped ,,align=center] {Start \\ Oven} (q_2)
    (q_1) edge  node  [midway,above, sloped,align=center] {Start  \\ Oven} (q_3);

\end{tikzpicture}
\caption{Microwave oven FSM without reachable counterexample}
\label{fig:UnReachableCE}
\end{figure}
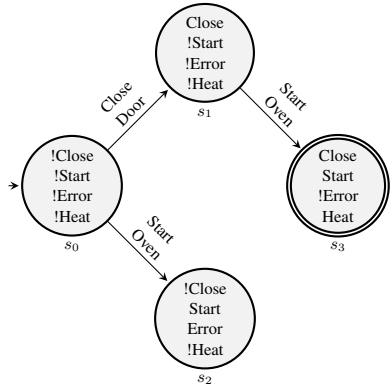

To illustrate this, we use a modified version of the microwave oven example, introduced in~\cite{clarke1999model}, as presented in \Cref{fig:UnReachableCE}. A safety requirement would be that the domain model does not allow the generation of erroneous plans, in LTL $p_0 = G(\neg \text{\em Error})$, where $G$ is the LTL {\em globally} operator.
Unconstrained verification will return $\langle StartOven \rangle$ as a counterexample that when applied to $s_0$ will produce $s_2$ which is an error state. However, when this model is used to find a plan that achieves the goal $(g = \text{\em Heat})$, this sequence will not be considered by the planner as it does not lead to a state that achieves the goal. 
Moreover, we observe that the valid plan $\langle CloseDoor,StartOven\rangle$ does satisfy the property $p_0$, i.e.\ is error-free. Thus, the sequence $\langle StartOven\rangle$ from $s_0$ to $s_2$ is an invalid planning counterexample; it does not achieve the goal, nor is it part of a valid plan towards the goal.

Counterexamples that are unreachable by planners, while searching for specific goals, exist in the literature. 
For example,~\citet{smith2005model} used the Spin model checker to verify whether a planning domain model would permit an automated planning system to select plans that would waste resources and therefore not meet the mission's science goals.   
To express this requirement, they used \textit{``five data-producing activities must be scheduled by any returned plan''} as a property for model checking. 
The automated system has two data-producing and two data-consuming activities, and a buffer that can hold four data blocks. The goal of the planner is to schedule five data-producing activity instances.
The counterexample returned by the model checker represented a plan with the two data-consuming activities scheduled before four data-producing activities. 
This plan did not contain a fifth data-producing task, because the data buffer was full after four data-producing activities and the only two data-consuming tasks that would have freed space in the buffer, were scheduled at the beginning of the plan with no data in the buffer. 
Though the model checker found a counterexample to falsify the property, we argue that any sound planner would not generate such a plan, because it does not achieve the planning goal.
As such, this counterexample would have been pruned during the planner's goal search, and consequently, it would never have been returned as a plan, i.e.\ it is unreachable for the planner, yet reachable by a goal-ignorant model checker. 
For this reason it constitutes an invalid planning counterexample.

The problem with invalid planning counterexamples is that they mislead the designer to unnecessarily restrict the domain model in the process of removing them. Consequently, debugging is made harder and genuine counterexamples could potentially be introduced in the process.
To overcome this, we observe that planning is performed for a specific goal. To exploit this observation for domain model verification, we propose to use the goal given to the planner as constraint to ensure that the counterexamples returned by a model checker, or other tools used in this context, falsify the given safety property while also achieving the planning goal. Thus, instead of performing unconstrained domain model verification, we propose goal-constrained verification of planning domain models. 
The details of this method are further explained in the next section.

\section{Goal-constrained verification of planning domain models}\label{Verification_planning_domain_models}
Planning domain model verification covers different objectives, including the domain's correctness, completeness, robustness, effectiveness and safety. 
The intent of safety verification in this context is to verify that any plan produced from the DUV will satisfy a given safety property. In other words, a domain is considered safe if the domain is guaranteed only to produce plans that satisfy the given safety property when used by a sound planner. This verification task can be performed using advanced search algorithms, such as model checkers or classical planners, to find a valid counterexample for the given safety property. 

We define a valid planning counterexample to be a sequence of actions that, firstly, can falsify the given safety property, secondly, can achieve the planning goal from the given initial state, and, thirdly, none of the sub-sequences of the counterexample can achieve the goal.  

Formally, this is defined as follows:
Let the planning problem, $P$, be a tuple, $(D,s_0, g)$, where $D$ is the domain model that describes the set of all available actions, $A$, $s_0$ is the initial state and $g$ is the desired goal. 
The plan $\pi$ is a solution to the planning problem $P$, defined as a sequence of actions, $\pi = \langle a_0,a_1,...,a_n \rangle$. These actions are chosen from $A$, $a_i \in A$, such that $\pi \models g$. In other words, when $\pi$ is applied to the initial state $s_0$ it yields a sequence of states $S(\pi)$, $S(\pi) = \langle s_0,s_1,...,s_n \rangle$ where the last state $s_n$ satisfies the planning goal $g$, $s_n \models g$. We say a plan $\pi$ satisfies a property $p$, $\pi \models p$, if the sequence of states $S(\pi)$, generated by the plan $\pi$, satisfies the property $p$, $S(\pi) \models p$.

Furthermore, as defined in~\cite{ghallab2004automated}, we call a plan $\pi$ a \textit{redundant plan}, if $\pi$ contains a subsequence, $\pi'$, $\pi'\ | \ \pi$, that achieves the goal $g$. 

\textbf{Definition 1:} 
A \textbf{valid planning counterexample} for a safety property, $p$, of a planning problem is a \textit{non-redundant plan}, $\pi$, that falsifies the safety property, $\pi \not\models p$.

Plans are required to be non-redundant in the definition of valid planning counterexamples to exclude any plans that are enriched with action sequences which are unnecessary to achieve the planning goal but required to falsify the given safety property.
Since sound planners can produce valid plans that have redundant subsequences, the scope of our method is limited to non-redundant planners i.e.\ planners that are guaranteed to produce non-redundant plans.

To ensure the returned counterexamples are valid, we constrain the verification problem with a goal, and we exclude any counterexample that is a redundant plan. 
More formally, the verification problem associated with planning task $P$ is defined as the tuple $V = (D, (s_0,g), p)$. Where $p$ is a formal safety property extracted from a given specification and required to hold over all valid paths that achieve the goal $g$ from the initial state $s_0$. 
It is important to highlight that although constraining the domain model verification with planning goals limits the verification results to planning problem instances, this is a necessary measure to obtain a verification method that is robust against invalid planning counterexamples.
We note that none of the current planning domain model verification methods verify planning domain models in their generality. 
Firstly, current methods require a grounded model, which represents a finite set of planning problems, in contrast to the infinite set of planning problems that non-ground domain models represent. 
Secondly, all methods need a specific initial state to be able to perform the verification tasks.
In our approach, by using the planning goal as one further constraint, we perform verification of single planning instances. 
This restriction is the cost associated with delivering a verification method that is robust against invalid planning counterexamples.

In this section, we introduced and formally defined the concept of goal-constrained verification of planning domain models. In the following subsections, we demonstrate how this concept can be realised using model checkers and state trajectory constraints planning techniques.

\subsection{Goal-constrained planning domain model verification using model checkers}

Model checkers verify safety properties by searching for counterexamples that falsify those properties. In the case of planning applications, any sequence of actions that does not achieve the given goal, will be pruned by any sound planner. Therefore, in the verification of planning problems, any counterexample that does not achieve the goal of the planning problem should be eliminated on the basis that this counterexample is unreachable by the planner.

To force model checkers to only return valid planning counterexamples, the safety property is first negated and then joined with the planning goal in a conjunction. This conjunction is then negated and supplied to the model checker as an input property. The final property requires the model checker to find a counterexample that both, falsifies the safety property and satisfies the planning goal. 
Note that, unlike Def.~1, this permits sequences that falsify the property after satisfying the goal. However, once the goal is achieved, planners terminate the search, thus rendering such sequences unreachable.
To eliminate these sequences, model transitions should be augmented with an additional guard, representing the negation of the goal, to restrict all transitions once the goal is achieved. With this modification, the model checker is forced to return counterexamples that falsify the safety property before achieving the goal, because once the goal is satisfied no further transitions will be permitted.

For a verification problem, $V = (D, (s_0,g), p)$, we first check whether the planning goal is achievable, then we translate the domain model $D$ into the model checker's input language, obtaining the model $M$ that incorporates the initial state $s_0$.
Then, the model $M$ is modified to $M'$ by augmenting the guards of all transitions with the negation of the goal condition.
From the definition of $M'$, we can derive two properties: First, \textit{\textbf{P1}: all plans generated from $M$ are also plans that can be generated from $M'$}.
Proof: any sequence of transitions from $M$ that ends with a transition that achieves the goal is also a sequence of transitions from $M'$. These sequences represent valid plans as they terminate with a transition that achieves the goal.
Therefore, all plans generated from $M$ are also plans in $M'$. 
Second, \textit{\textbf{P2}: any valid counterexample for $M'$ is also a valid counterexample for $M$}.
Proof: as $M'$ is a more constrained version of $M$, the set of all legal transition of $M'$, $\Pi(M')$, is contained in the set of all legal transitions of $M$, $\Pi(M)$, 
i.e.\ $\Pi(M) \supseteq \Pi(M')$. 
It follows that any valid counterexample in $\Pi(M')$ is also in $\Pi(M)$.

The model checker is then applied to the verification problem $V'' = (M', p')$, where $p'$ is defined using $F$, the LTL {\em eventually} operator~\cite{clarke1999model}, as follows:
\begin{alignat}{1}
p' = \neg \big( F(\neg p) \wedge F(g) \big)  \label{eq-inpuProperty}
\end{alignat}
There are two possible outcomes.
If the model checker returns a counterexample $\pi$:
\begin{alignat}{1}
 &\exists \pi \in \Pi(M'). \pi\not\models p' \\
&\equiv\ \exists \pi \in \Pi(M'). \pi \models (F(\neg p) \wedge F(g)) 
\end{alignat} 
From the definition of the LTL {\em eventually} operator $F$:
\begin{alignat}{1}
\exists i \ge 0, \ s_i \in S(\pi), s_i & \models \neg p \label{eq-property1} \\
\exists j \ge 0, \ s_j \in S(\pi), s_j & \models g \label{eq-goal1} 
\end{alignat}  

It follows that there is at least one sequence $S(\pi)$ that falsifies the property $p$, and there is a state $s_j$ in that sequence which satisfies the goal $g$, according to (\ref{eq-property1}) and (\ref{eq-goal1}).
In addition to that, in the sequence $S(\pi)$, $p$ is guaranteed to be falsified before $g$ is satisfied. This is because $\pi \in \Pi(M')$ and $M'$ is constrained to not produce any transitions after achieving the goal. Thus, the counterexample $\pi$ is a valid planning counterexample in $M'$ for the original safety property $p$ as per Def.~1. 
Furthermore, from (\textit{\textbf{P2}}), $\pi$ is also a counterexample in $M$. This proves that the DUV does not satisfy the safety property $p$ {\em with respect to the goal g.}

The other potential outcome is that the model checker fails to find a counterexample, then $\forall \pi \in \Pi(M')$:
\begin{alignat}{1}
&\ \pi \models p' \\
&\equiv\ \ \pi \models \neg ( F(g) \wedge F(\neg p)) \\
&\equiv\ \ \pi \models (\neg F(g) \vee \neg F(\neg p)) \\ 
&\equiv\ \ \pi \models \neg F(g) \vee \pi \models \neg F(\neg p) \\ 
 &\equiv\ \ \pi \not\models F(g) \vee \pi \models G(p) \\
 &\equiv\ \ \pi \models F(g) \Rightarrow \pi \models G(p) \label{eq-modelSafePlan}
\end{alignat}

Where $G$ is the LTL {\em always} operator. This means $p$ is always true for any sequence of actions in $M'$ that achieves the goal, i.e.\ for all possible plans.
Since from (\textit{\textbf{P1}}) all plans generated from $M$ are also plans in $M'$, and from (\ref{eq-modelSafePlan}) all plans in $M'$ are safe, we can conclude that all plans in $M$ are safe.
This proves that the DUV satisfies the original property {\em with respect to the goal.}
\subsection{Goal-constrained planning domain model verification using planning techniques} \label{verification_using_planning}

Domain models can be verified to only produce valid plans, in terms of satisfying a given safety property, for a specific goal using planners that use breadth first search.
This is achieved by consulting the planner over the DUV to produce a plan that can satisfy the goal and the negation of the property.
If the domain model permits producing plans that, along with achieving the goal, contradict the safety property, then an unsafe plan can be found.
Thus, the returned plan is a counterexample that demonstrates that the safety property does not hold.
On the other hand, if the domain model does not permit the generation of plans that can satisfy the negation of the safety property while achieving the goal, then the planner will fail. Thus, the  property holds in any plan produced for the given goal. 
The following subsection provides a description of how state trajectory constraints can be used to verify planning domain models for a specific goal.
\subsubsection{Goal-constrained planning domain verification using planning techniques with state trajectory constraints}

The PDDL3.0 state trajectory constraints~\cite{gerevini2006preferences} can be used to perform planning domain model verification.
First, the negation of the given property is expressed using PDDL3.0 modal operators and embedded in the original domain model as state trajectory constraint.
The modified model is then used by a planner, as described earlier, to perform the verification.

or a verification problem, $V = (D, (s_0,g), p)$, the safety property, $p$, is negated and expressed in terms of PDDL3.0 modal operators as shown in~\cite{gerevini2009deterministic}.
The result is added as a state trajectory constraint to the original domain model.

Using the algorithm proposed in~\cite{edelkamp2006costoptimal}, the new model is transformed into a PDDL2 compatible version.
This is performed by first translating the state trajectory constraint into a non-deterministic finite state automaton (NFA). The NFA which can capture property violations is then embodied in the model in terms of additional predicates and conditional effects.
These additions observe the behaviour of the automaton that represents the constraint and stop goal satisfaction unless the constraint is satisfied too.
This yields a new planning problem, $P' = (D',s_0', g')$, where $D',\ s_0',\ g'$ are modified instances of $D,\ s_0,\ g$ that are supplemented with the additional predicates and conditional effects of the automaton that represents the introduced constraint.
Let the set of legal sequences of actions that can be generated from $D$ be $\Pi(D)$ and from $D'$ be $\Pi(D')$.
Note that $D'$ is an augmented version of $D$ and the additions to $D'$ do not affect the number of original operators, their preconditions, or their effects. Furthermore, the additional conditional effects do not affect the original predicates. Hence, $\Pi(D) = \Pi(D')$.

Then, a planner is applied to $P'$ with two possible outcomes. If the planner finds a plan $\pi$ then: $\exists \pi \in \Pi(D').\ \pi \models g'$.
Since the satisfaction of $g'$ implies both, the satisfaction of the original goal $g$ at the last state of the sequence $S(\pi)$, and the satisfaction of the state trajectory constraint (the negation of the safety property) by the sequence $S(\pi)$: $\exists \pi  \in \Pi(D'). \ (\pi \models g \wedge \pi \models \neg p)$. Furthermore, since $\Pi(D) = \Pi(D')$:
\begin{alignat}{1}
\exists \pi  \in \Pi(D). \ (\pi \models g \wedge \pi \models \neg p)\label{eq-constraint2}
\end{alignat}
Furthermore, from (\ref{eq-constraint2}) it follows that $\pi \not\models p$,
confirming that there is at least one sequence of actions from $D$ that achieves the goal while not respecting the safety property. 
Therefore, this sequence is a valid planning counterexample for that property as per Def.~1. Hence, the DUV does not satisfy the property {\em wrt.\ the planning goal.}
Alternatively, if the planner fails to find a plan, then, as opposed to (\ref{eq-constraint2}), we have:
\begin{alignat}{2}
& \nexists \pi \in \Pi(D). \ (\pi \models g \wedge \pi \models \neg p) \\
&\equiv \forall \pi\in \Pi(D). \ \neg(\pi \models g \wedge \pi \models \neg p) \\
&\equiv \forall \pi\in \Pi(D). \ \neg(\pi \models g \wedge \neg (\pi \models p)) \\
&\equiv \forall \pi \in \Pi(D). \ (\neg (\pi \models g) \vee  \pi \models p) \\  
& \equiv \forall \pi \in \Pi(D). \ (\pi\models g \Rightarrow \pi\models p) \label{eq-last}
\end{alignat}

Hence, any sequence of actions from $D$ that achieves the goal also satisfies the safety property. Therefore, the property holds for the planning domain model {\em wrt.\ the given goal.}

\section{Example } \label{Example_CaveDiving}
In this section, we discuss how goal-constrained planning domain verification can verify safety properties using both the Spin model checker and the MIPS-XXL planner with breadth first search option.
We perform constrained and unconstrained verification tasks to show how unlike the latter task our method does not return unreachable counterexamples.
As an example, we consider the classical cave diving planning domain taken from the Satisficing Track of the IPC-2014~\cite{IPC2014}.
The problem consists of an underwater cave system with a finite number of partially interconnected locations.
Divers can enter the cave from a specific location, entrance, and swim from one location to an adjacent connected one.
They can hold up to four oxygen tanks and they consume one for every swim and take-photo action.
Only one diver can be in the cave at a time.
Finally, divers have to perform a decompression manoeuvre to go to the surface and this can be done only at the entrance.
Additionally, divers can drop tanks in or take tanks from any location if they hold at least one tank or there is one tank available at the location, respectively.

The planning goals of this domain, as provided in the problem files in the IPC-2014, consist of two parts.
The first part dictates the required underwater location of which the photo is to be taken (we call it mission target) and the second part which mandates the divers should return to the surface after completing the mission (we call it safety target). 

A critical safety property, $p$, is that divers should not drown i.e.\ they should not be in an underwater location, other than the entrance, where neither the divers nor the location has one full oxygen tank at least.

To enable the planner and the model checkers to explore the entire state space, we considered only one diver and we modified some actions to enable the diver to go back into the water after a dive.
These modifications are further explained in the commented simplified planning domain model PDDL file which is provided along with the tasks problem PDDL and Promela files online \footnote{\url{https://github.com/Anas-Shrinah/Goal-constrained-planning-domain-model-verification-repository}}.

First, we translated the planning domain model from PDDL to Promela. Thus, the verification results using the translated model only hold provided that the translation is valid. The verification of the translation is outside the scope and focus of this paper and left for future work.

In this example, the chosen planning goal is to have a photo of the first location, $L_1$, and to get the diver outside the water. The verification tasks are:

1 - Unconstrained verification with only the safety property $p$: Both Spin and MIPS-XXL found a counterexample could be $\langle$\textit{prepare-a-tank, enter-water, swim}$(L_0,L_1)\rangle$. Indeed, this counterexample leads the diver to a drowning state.
At the end of this sequence, the diver will have consumed their oxygen tank and will be in underwater location $L_1$. This is not the entrance, so they can not surface and they do not have an oxygen tank to swim back to the entrance.
However, this is not a plan because it does not achieve any useful goal. Therefore, it will not be produced by any sound planner when it is used in a practical scenario (taking a photo of any location).

2- Verification with safety property and incomplete goal (mission target only): Both Spin and MIPS-XXL returned
$\langle$\textit{prepare-tank, prepare-tank, enter-water, swim}$(L_0,L_1)$\textit{, take-photo}$\rangle$.
This counterexample achieves the goal and violates the property.
% %
However, without the safety part of the goal, it would be possible to generate plans that imply divers should swim to an underwater location and take a photo of it without requiring the divers to return to the surface. These kind of plans are illegal as they do not respect the safety part of the goal. Therefore, such sequences are unreachable counterexamples i.e.\ will never be produced by any sound planner while planning for a legal goal.

3- Verification using Spin with both safety property and proper goal but without the augmented model $M'$: Spin found a counterexample $\langle$\textit{prepare-tank, prepare-tank, prepare-tank, prepare-tank, enter-water, swim}$(L_0,L_1)$\textit{, take-photo, swim}$(L_1,L_0)$\textit{, decompress, enter water, swim}$(L_0,L_1)\rangle$. This counterexample ach\-ieves the goal and violates the safety property but only after the goal is achieved. Therefore, this is also an unreachable counterexample because a non-redundant planner will terminate after achieving the goal and any counterexample that violates the property after achieving the goal will not be returned. Hence, it is unreachable.

4- Goal-constrained planning domain verification, as presented in this paper. The result was: No plan is returned by the planner MIPS-XXL with complete exploration and no counterexample is returned by Spin with exhaustive verification mode. This means the planning domain model has no provision of producing a plan that can violate the safety property before achieving the goal, i.e.\ this model is safe with respect to the given property and goal pair.

Though the counterexamples returned by the incomplete verification tasks number one, two and three are obviously unreachable and should not misguide the designers to overcomplicate the model, in a real world sized application such invalid planning counterexamples can be critical and much more difficult to recognise and avoid. We expect that our proposed concept can save practitioners a huge amount of person-hours trying to alter planning domain models for behaviours that their planners will never experience in practice.

\begin{figure*}[h!]
 \centering
\begin{subfigure}[b]{.6\textwidth}
  \includegraphics[width=\linewidth]{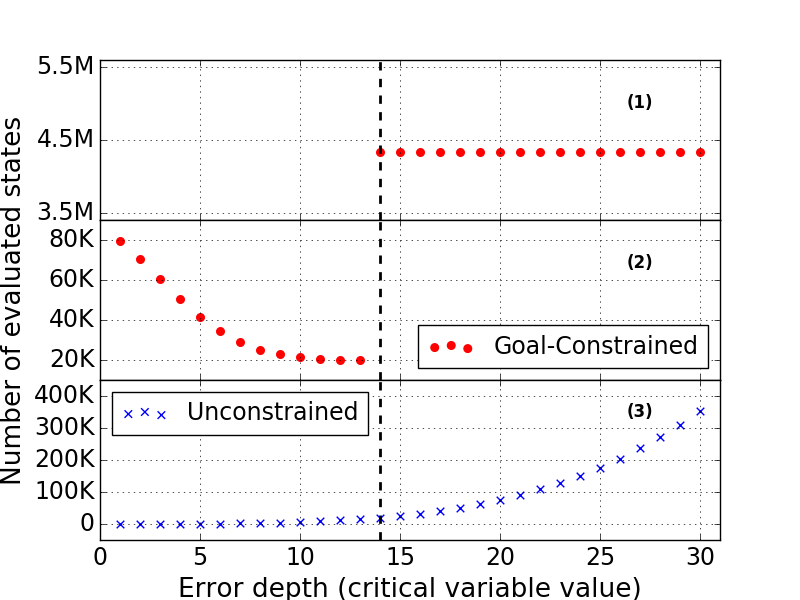}
  \caption{Evaluated states vs error depth\\ using Spin}
  \label{fig:EvaluatedStatesVsErrorDepthSPIN}
\end{subfigure}%
\begin{subfigure}[b]{.6\textwidth}
  \centering
  \includegraphics[width=\linewidth]{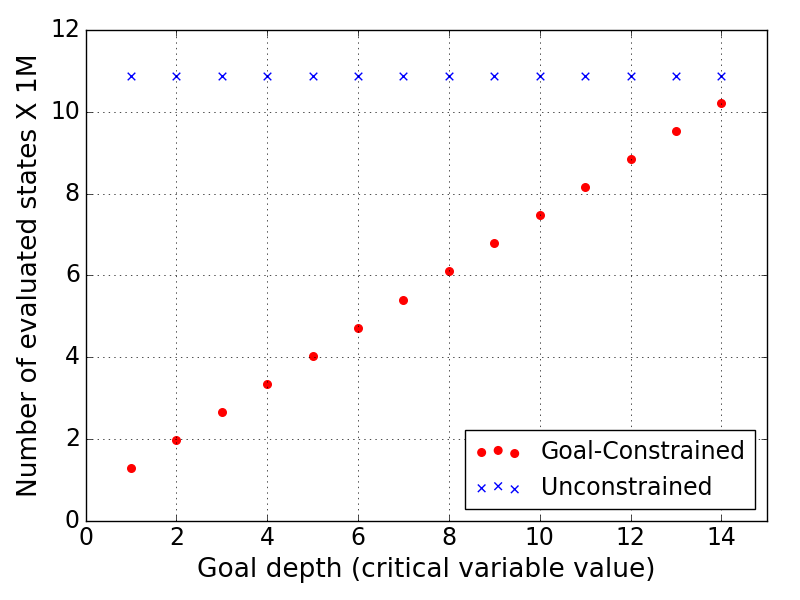}
  \caption{Evaluated states vs goal depth\\ using Spin}
  \label{fig:EvaluatedStatesVsGoalDepthSPIN}
\end{subfigure}
\begin{subfigure}[b]{.6\textwidth}
  \includegraphics[width=\linewidth]{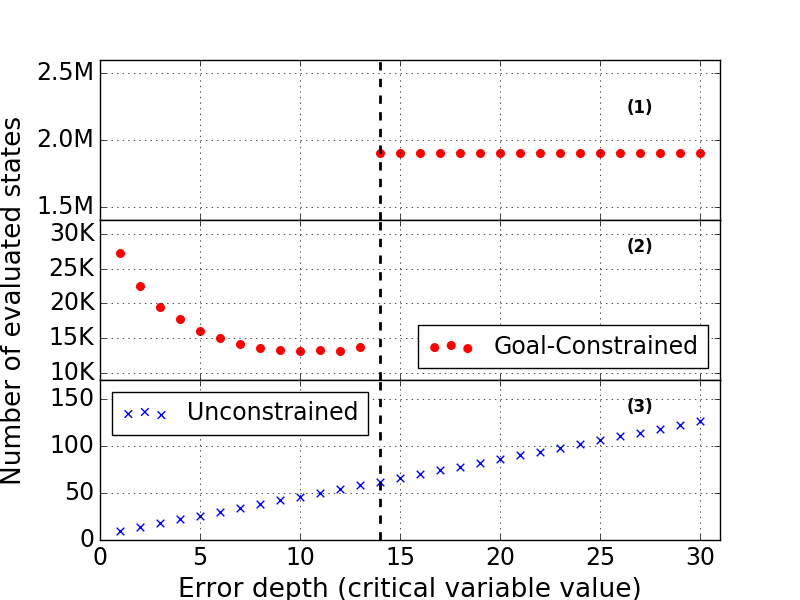}
  \caption{Evaluated states vs error depth\\ using MIPS-XXL}
  \label{fig:EvaluatedStatesVsErrorDepthMIPS}
\end{subfigure}%
\begin{subfigure}[b]{.6\textwidth}
  \centering
  \includegraphics[width=\linewidth]{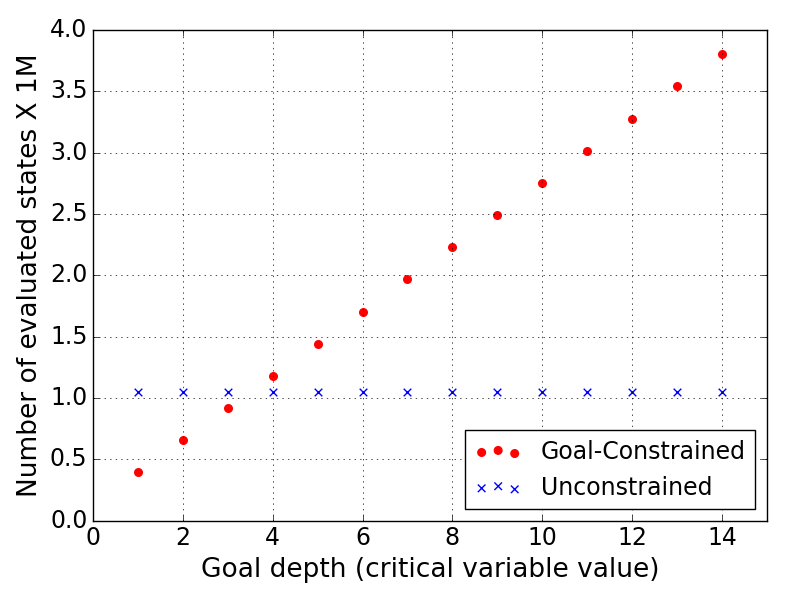}
  \caption{Evaluated states vs goal depth\\ using MIPS-XXL}
  \label{fig:EvaluatedStatesVsGoalDepthMIPS}
\end{subfigure}
\caption{The behaviour of our verification method with different verification tasks using Spin (a,b) and MIPS-XXL-IPC5 (c,d).}
\label{fig:figSpinAndMIPS}
\end{figure*}

\section{Experiments} \label{Experiments}

To evaluate the feasibility and the behaviour of our approach, we designed two experiments to investigate how constraining the verification with the planning goal impacts the verification cost. This cost is measured by the number of states evaluated by the verification tools to confirm whether or not a counterexample exists. We consider the number of evaluated states to be an objective measure that is repeatable on any hardware platform, as opposed to measuring execution time. The scripts to repeat the experiments along with the data are available online \footnote{\url{https://github.com/Anas-Shrinah/Goal-constrained-planning-domain-model-verification-repository}}.

\subsection{Experiment setup} \label{Experiment-setup}
The first experiment focuses on comparing the cost of both unconstrained and goal-constrained verification tasks while varying the safety property violation depth 
in order to explore situations with and without a valid planning counterexample. The safety property violation depth is hereafter termed ``error depth''. We synthesised a fully reachable model that consists of one critical and three independent variables, each with a range from 0 to 31. Each variable has two actions, one to increase and one to decrease its value by one. The goal is achieved when the critical variable value reaches 14. The error condition is changed from the value of the critical variable being 1 to 31. The range of the variables is chosen as 31 to expose any possible trends. Consequently, the number of variables is set to four to allow the model to be explored within a memory limit of 10 GB.

The second experiment investigates the effect of the early termination of the verification process, after achieving the goal, on the cost of verification tasks while increasing the depth of the planning goal. The model used in this experiment has one critical and four independent variables, each with a range from 0 to 15. Variables have two actions as in the previous model. This time, there is no error in the model and the goal condition is varied from critical value 1 to 14. The variables' range is reduced to 15 to permit increasing the number of variables to five while keeping the required memory within the 10 GB constraint.
Both experiments are performed using the Spin model checker and the MIPS-XXL-IPC5 planner with breadth first search option.

\subsection{Results and discussion} \label{Result-and-discussions}

The results of applying our approach in comparison to the unconstrained verification methods are as follows. The states evaluated by Spin and MIPS-XXL-IPC5 are presented in \Cref{fig:figSpinAndMIPS}.
In the first experiment, our approach showed broadly similar behaviour when it was applied using Spin in \Cref{fig:EvaluatedStatesVsErrorDepthSPIN} and MIPS-XXL-IPC5 in \Cref{fig:EvaluatedStatesVsErrorDepthMIPS}. 
Note that the aim of these experiments is to showcase the feasibility of using our approach and to explore its behaviour, rather than comparing the performance of the verification tools. We believe such comparison depends heavily on the model under verification, for more insights the reader is referred to~\cite{edelkamp2003limits,Albarghouthi09onthe,li2012planning}.
Ergo, we focus our discussion on the results obtained from model checking (\Cref{fig:EvaluatedStatesVsErrorDepthSPIN}).

The vertical line in \Cref{fig:EvaluatedStatesVsErrorDepthSPIN} marks the goal level (critical value of 14) and splits the graph into two areas.
On the right-hand side, the errors are deeper than the goal, i.e.\ the errors can only be reached after the goal is achieved. Thus, these errors are regarded as invalid planning counterexamples by our method as per Def.~1.
Therefore, unlike unconstrained verification approaches, our method continues its exhaustive search to confirm the non-existence of any valid planning counterexamples. Thus, our method evaluates the maximum number of states for these verification tasks as shown in \Cref{fig:EvaluatedStatesVsErrorDepthSPIN}-(1).

On the left-hand side, the errors are shallower than the goal, i.e.\ the errors are reachable before achieving the goal. Hence, these errors are considered as valid planning counterexamples according to Def.~1.
For the same verification task, \Cref{fig:EvaluatedStatesVsErrorDepthSPIN}-(2) shows that our method assesses more states than the unconstrained approaches as depicted in \Cref{fig:EvaluatedStatesVsErrorDepthSPIN}-(3).
This is due to the fact that after finding an error, a safety property violation, our method keeps exploring and searching for a path to the planning goal while traditional methods terminate as soon as an error is found.
However, the short counterexamples returned by these methods may or may not be valid planning counterexamples, whereas our method is guaranteed to return valid planning counterexamples only. The extra states visited by our approach are the cost associated with this guarantee.

In \Cref{fig:EvaluatedStatesVsErrorDepthSPIN}-(2) (and in \Cref{fig:EvaluatedStatesVsErrorDepthMIPS}-(2), respectively), we notice a drop in the number of evaluated states by our method as the error depth gets closer to the goal depth.
This is attributed to the fact that the safety property (state trajectory constraint) in the model checker (planner) is translated into an automaton. This automaton influences the state space exploration during the verification process.
The automaton has a transition that is activated when an error is reached. Therefore, if an error is reached in an early stage in the verification, the error transition is triggered and the verification tool is forced to explore more states than if the error transition was triggered closer to the goal.
Once both the error and the goal transitions are triggered, then the automaton reaches an acceptance state. 
Thus, the search terminates with a valid planning counterexample.

In the second experiment, Figure~\ref{fig:EvaluatedStatesVsGoalDepthSPIN} shows that when using Spin with a planing domain model with no counterexample, our approach explores fewer states than unconstrained verification methods.
This reduction in the verification cost is caused by the early termination of the verification search once the goal is achieved and no error could be found at shallower depths. This advantage of the goal-constrained verification approach comes at the cost of limiting the verification results to a single planning goal.
Additionally, it is observed that the number of evaluated states by the goal-constrained verification method rises as an effect of the increasing goal depth. This is caused by the expansion of the part of the model that needs to be checked as the goal depth increases. On the other hand, the unconstrained methods visit a constant number of states as they are independent from the goal depth by definition.

Another interesting observation when using the planner in Figure~\ref{fig:EvaluatedStatesVsGoalDepthMIPS} is that our method explores more states than the unconstrained verification approaches when the planning goal depth is more than three. This behaviour is caused by the interaction of two factors. 
In our approach, MIPS-XXL-IPC5 translates the state trajectory constraint to an automaton which is then incorporated in the planning domain model. Thus, the model used in our method is more complicated than the model used by the unconstrained approaches were state trajectory constraints are not used. After a certain depth of the planning goal, the extra states evaluated as a result of the additional state trajectory constraint in our method outweigh the saving from the early termination of the verification process.

\section{Conclusions and future work } \label{conclusion}

The verification of planning domain models is essential to guarantee the safety of planning-based automated systems. 
Invalid planning counterexamples returned by unconstrained planning domain model verification techniques undermine the verification results. They can mislead system designers to perform unnecessary remediations that can be prone to errors.
In this paper, we introduced goal-constrained verification, a new concept to address this problem, which restricts the verification task to a specific goal. 
This limits counterexamples to those practically reachable by a planner that is tasked with achieving the goal. 
Consequently, our method verifies the domain model only wrt.\ a specific goal. We consider this to be an acceptable limitation, given that planners also operate on this basis.
Since we have excluded redundant plans from our definition of valid planning counterexamples, the verification results of our method are limited to the application of non-redundant planners. A weaker form of non-redundancy will be considered in future work.

We have demonstrated how model checkers and planning techniques can be used to perform goal-constrained planning domain model verification.
Our experimental evaluation confirms the feasibility of our method and presents its benefits and limitations compared to unconstrained verification methods. 
The proposed technique is simple which makes it readily usable in practice. It is also effective as formally proven in the paper.

We note that a grounded planning domain model defines a finite set of planning problems. For our method to completely verify such a set, it has to be repeatedly applied to every planning goal. In practice, we have noticed that only a small number of predicates is typically used to specify the planning goals. Thus, we expect that in applications where the set of planning goals is relatively small, our method could exhaustively verify the complete set of planning problems, especially if the tools take advantage of the latest optimisation techniques to reduce computational complexity.

As future work we intend to employ verification techniques such as verification reusability and compositionality~\cite{10.1007/3-540-49059-0_12}, abstraction~\cite{clarke1994model} and symmetry reduction~\cite{clarke1998symmetry} to perform complete verification for grounded planning domain models.

\section*{Acknowledgements}
The authors are grateful to Derek Long for his useful comments.

\bibliographystyle{named}
\bibliography{Verification-of-Planning-Domain-Models-ICAPS-2020-bibliography}

\end{document}